\documentclass[letterpaper, 10 pt, conference]{ieeeconf}  
\usepackage{mathtools}
\usepackage{ifluatex,ifxetex}
\ifluatex
  \usepackage{fontspec} 
\else\ifxetex
  \usepackage{fontspec} 
\else 
  \usepackage[T1]{fontenc}
  \usepackage[utf8]{inputenc} 
\fi\fi

\IEEEoverridecommandlockouts                              

\usepackage{amssymb}
\usepackage{color}
\let\labelindent\relax
\usepackage[inline]{enumitem}
\usepackage{times}
\usepackage{epsfig}
\usepackage{bm}
\usepackage{graphicx}
\usepackage{amsmath}
\usepackage{amssymb}
\usepackage[font=small]{caption}
\usepackage{floatrow}
\usepackage{mathrsfs}
\usepackage{subcaption}
\usepackage{floatrow}
\usepackage{xcolor,colortbl,multirow,hhline}
\usepackage{array}
\usepackage{wrapfig}
\usepackage{booktabs}
\usepackage{tabularx}

\definecolor{LightCyan}{rgb}{0.88,1,1}
\definecolor{lightgray}{HTML}{EEEEEE}
\definecolor{oneGray}{gray}{0.95}
\definecolor{twoGray}{gray}{0.9}
\definecolor{threeGray}{gray}{0.85}
\newcolumntype{a}{>{\columncolor{oneGray}}c}
\newcolumntype{b}{>{\columncolor{twoGray}}c}
\newcolumntype{d}{>{\columncolor{threeGray}}c}
\usepackage[pagebackref=true,breaklinks=true,colorlinks,bookmarks=false]{hyperref}

\begin{document}





\title{\LARGE \bf MATRIX: Multi-Agent Trajectory Generation with Diverse Contexts}

\author{Zhuo Xu*$^{1}$, Rui Zhou*$^{2}$, Yida Yin*$^{3}$, Huidong Gao$^{3}$, Masayoshi Tomizuka$^{3}$, and Jiachen Li$^{4}$
\thanks{*indicates equal contribution.}
\thanks{$^{1}$Everyday Robots, X, USA {\tt\small
zhuoxu@google.com}}
\thanks{$^{2}$Massachusetts Institute of Technology, USA {\tt\small zhourui@mit.edu}}
\thanks{$^{3}$University of California, Berkeley, USA {\tt\small \{davidyinyida0609, hgao, tomizuka\}@berkeley.edu}}
\thanks{$^{4}$University of California, Riverside, USA {\tt\small jiachen.li@ucr.edu}} 
}
\maketitle

\thispagestyle{empty}
\pagestyle{empty}

\begin{abstract}
Data-driven methods have great advantages in modeling complicated human behavioral dynamics and dealing with many human-robot interaction applications. However, collecting massive and annotated real-world human datasets has been a laborious task, especially for highly interactive scenarios. On the other hand, algorithmic data generation methods are usually limited by their model capacities, making them unable to offer realistic and diverse data needed by various application users. In this work, we study trajectory-level data generation for multi-human or human-robot interaction scenarios and propose a learning-based automatic trajectory generation model, which we call Multi-Agent TRajectory generation with dIverse conteXts (MATRIX). 
MATRIX is capable of generating interactive human behaviors in realistic diverse contexts. We achieve this goal by modeling the explicit and interpretable objectives so that MATRIX can generate human motions based on diverse destinations and heterogeneous behaviors. We carried out extensive comparison and ablation studies to illustrate the effectiveness of our approach across various metrics. We also presented experiments that demonstrate the capability of MATRIX to serve as data augmentation for imitation-based motion planning.
\end{abstract}
\section{Introduction}
Interactive human behavioral dynamics are among the most challenging dynamics to model due to the complicated hidden features and the diverse behaviors. Researchers have recently favored data-driven methods as the solution for a wide range of human-robot interactive problems. Nevertheless, collecting real-world human motion datasets even only on the trajectory level is not an easy task because it can take huge amounts of human volunteer recruiting or label annotation efforts. Although some algorithmic methods can generate a variety of human trajectory data based on deterministic or stochastic algorithmic motion generators \cite{van2008reciprocal, van2011reciprocal}, most of these rule-based approaches can only do well in limited domains and yet fail to produce realistic and smooth trajectory data for general purposes. While the demand for massive human behavior datasets is further exacerbated by the advance of deep learning-based methods, the plenty of available trajectory data can also benefit the learning of the human\begin{figure}[t]
\includegraphics[width=\linewidth]{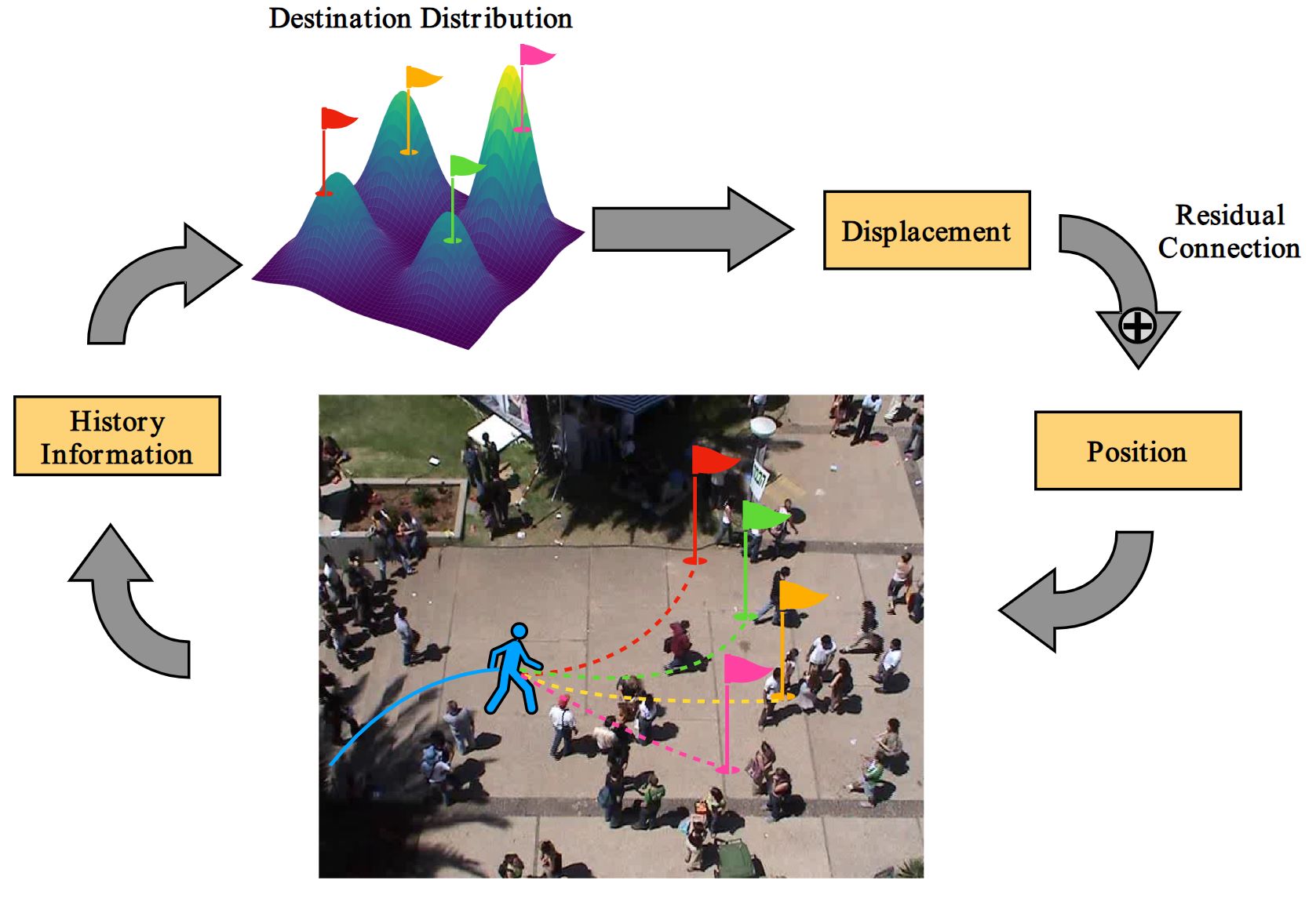}

\caption{
\textbf{Illustration of the MATRIX generation process.} Based on the observed trajectory and sampled destinations, MATRIX can generate various heterogeneous trajectories through residual connection.}

\label{fig:teaser}
\end{figure} motion logic, which in turn assists assorted downstream tasks with human-robot interaction. To this end, equipped with deep neural networks, researchers have built a variety of generative models by encoding highly multi-modal and uncertain human motions into latent states \cite{salzmann2020trajectron++, choi2021shared, zhou2022grouptron, dax2023disentangled}. However, these models are based on the assumption that the underlying randomness in human behaviors is a conditional Gaussian distribution. Thus, their performances depend on how well the latent space models the hidden features of agents, and the diversity of the generated data cannot be guaranteed.

To achieve explainable learning of the human behavioral latent features, we enable the human trajectory generative model to produce diverse and distinct contexts by explicitly modeling one of the most prominent properties that affect human behaviors: the temporal traveling destinations. We call our model Multi-Agent TRajectory generation with dIverse conteXts (MATRIX), which adopts a conditional variational autoencoder framework and self-supervised training scheme while featuring a Gaussian Mixture Model (GMM) for modeling the hidden distribution of temporal destinations, which naturally exhibits multi-modality, meaning diverse future interaction modes can emerge from same past trajectory contexts. In addition, GMM provides explainable parameters that we can regularize throughout training to forfeit mode collapse and guarantee diverse generated behaviors. As a realistic human trajectory data generator, MATRIX also enforces soft safety constraints by adopting residual structures that produce human actions. 
The illustration of the trajectory generation process is shown in Fig. \ref{fig:teaser}.
We evaluated MATRIX in the human crowd navigation setting, against a variety of baseline approaches and ablation models, with a series of metrics demonstrating MATRIX's capability of generating realistic and diverse trajectory data. The main contributions of this paper are as follows:

\begin{enumerate}
    \item We present a novel human trajectory generation framework called MATRIX, which produces diverse and realistic human motion data.
    \item We design a GMM to explicitly model the distribution of human temporal destinations and utilize residual action to control the aggressiveness of sampled trajectories and encourage diverse generated behaviors.
    \item We introduce several novel motion primitive distribution shifts (Chi-square distance $\bm{\chi^2}$ between real and generated trajectories) as the realism metrics in addition to the classic waypoint displacement error metrics.
    \item Our approach, as evaluated against a series of baselines and ablations, achieves state-of-the-art performance as a diverse trajectory data generator, in terms of quantitative diversity and realism metrics on the ETH/UCY benchmarks, and as demonstrated in experiments where it serves as a data source for the training set augmentation for a downstream behavior cloning task.
\end{enumerate}
\section{Related Work}
\subsection{Human Trajectory Prediction and Planning}
Data-driven navigation behavior, interaction understanding, prediction and planning have garnered significant attention from numerous communities in recent years \cite{li2023game, chang2020cascade, xu2019toward, xu2020guided, chen2020end, li2020evolvegraph, xu2021history,li2024multi,mahadevan2024generative,nasiriany2024pivot}. 
Some of the earlier works in human trajectory forecasting and planning include the social force model \cite{helbing1995social}, the dynamic potential field \cite{treuille2006continuum}, velocity-based collision avoidance \cite{van2011reciprocal}, and model predictive control \cite{sun2023distributed}. Machine learning has also been utilized to forecast human trajectories, by modeling it as a deterministic time-series regression problem and solving it using Gaussian Process Regression (GPR) \cite{rasmussen2003gaussian}, inverse reinforcement learning (IRL) \cite{lee2016predicting}, and recurrent neural networks (RNNs) \cite{ morton2016analysis, vemula2018social}. 
More recent generative approaches have adopted a recurrent architecture with a latent space, such as a conditional variational auto-encoder (CVAE) \cite{deo2018multi, li2019conditional, salzmann2020trajectron++, girase2021loki, lee2017desire, chen2021human,li2021spatio,dax2023disentangled}, a generative adversarial network (GAN) \cite{gupta2018social, sadeghian2019sophie,li2019interaction}, or a diffusion model \cite{gu2022stochastic, jiang2023motiondiffuser, barquero2023belfusion, mao2023leapfrog} to encode multi-modality. 
Some other works also focus on improving the stochastic process for multi-modal prediction \cite{bae2022non}. On the modeling side, Graph Convolutional Networks (GCN) are first introduced in \cite{kipf2016semi}, and Spatio-Temporal Graph Convolutional Networks (STGCN) \cite{yan2018spatial} and Social-STGNN \cite{mohamed2020social} are designed to capture both spatial and temporal information.

There are also methods for human trajectory prediction that employ attention mechanisms \cite{li2021spatio,yuan2021agentformer,sun2022interaction,gu2022stochastic,li2021rain, HU2023109592}. AgentFormer \cite{yuan2021agentformer} employs an agent-aware attention mechanism. MID \cite{gu2022stochastic} is a Transformer-based framework with motion indeterminacy diffusion. ScePT \cite{chen2022scept} generates scene-consistent joint trajectory predictions with a tunable risk measure. Y-net \cite{Y-net} utilizes encoder and decoder architecture to reconstruct the heatmap of future trajectories. \cite{chen2019crowd} introduces a socially attentive network that consists of an interactive module that encodes interactions through local maps. \cite{chen2020relational} uses model-based deep reinforcement learning to plan actions. 
 
Compared to previous work, our method has better performance when evaluated as a data generator. The diverse trajectories generated by MATRIX show its potential for the downstream task and as an alternative to traditional data augmentation methods, such as translating and rotating.

\subsection{Multi-modality Encoding}
A key in predicting human behavior is encoding the multi-modality nature \cite{gu2022stochastic, ma2022multi}. Many works model agents' future modes implicitly as latent variables, including DESIRE \cite{lee2017desire} utilizing a conditional variational auto-encoder to obtain a diverse set of future prediction samples. PRECOG \cite{rhinehart2019precog} uses a flow-based generative model to perform both standard forecasting and the novel task of conditional forecasting. SocialGAN \cite{gupta2018social} further trains a network adversarially against a recurrent discriminator, and encourage diverse predictions with a novel variety loss. Some other works choose to discretize the output space as goals or anchors, then do predictions based on each goal or anchor, including MultiPath \cite{chai2019multipath} and Covernet \cite{phan2020covernet}. TNT \cite{zhao2021tnt} and DenseTNT \cite{gu2021densetnt} also first predict goal candidates and then generate separate trajectories conditioned on targets. In comparison, we utilize a Gaussian Mixture Model (GMM) to capture the goal of continuous distribution. This reduces computation load and improves the performance of goal estimation without relying on the quality of predefined goal anchors.
\begin{figure*}[!tbp]
\centering
\includegraphics[width=0.9\textwidth]{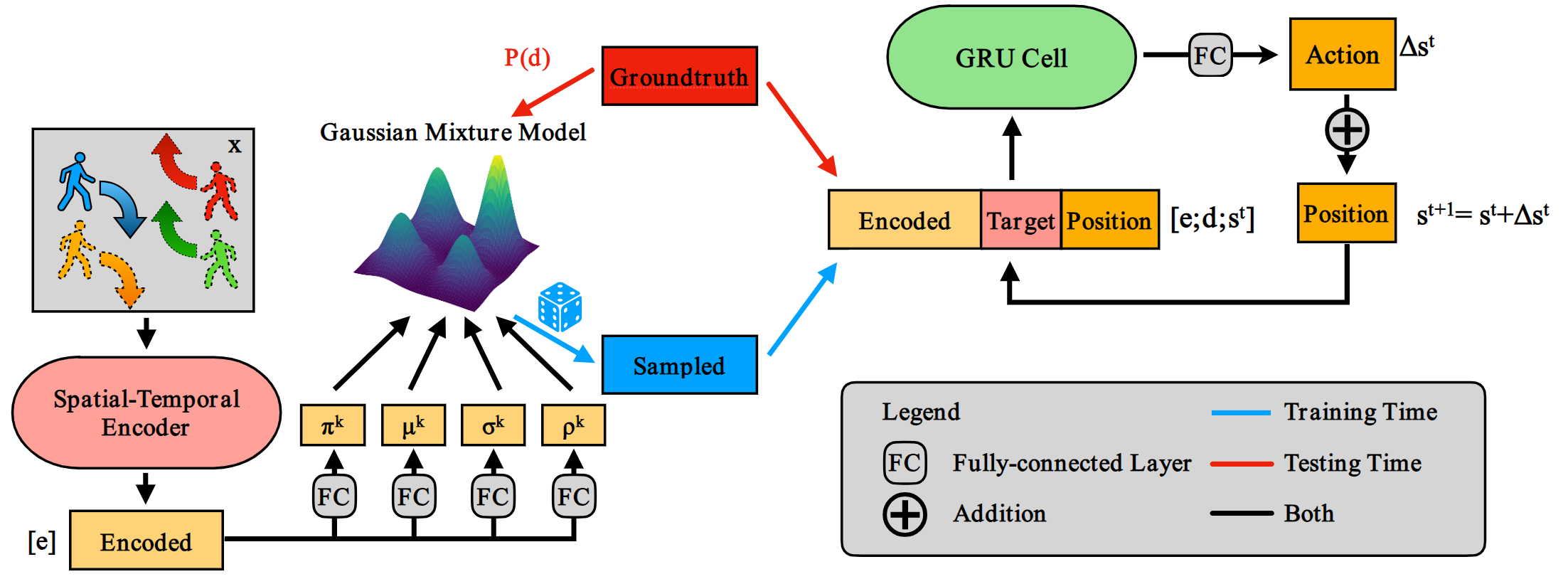}
\caption{\textbf{The architecture of MATRIX.} MATRIX consists of a Spatial-Temporal Encoder, a Gaussian Mixture Model (GMM), and a Gated Recurrent Unit (GRU) Decoder with a residual layer.}

\label{fig:model}
\end{figure*}
\section{MATRIX}
In this section, we introduce our approach MATRIX with a focus on encouraging the generation of diverse trajectories. At each time step $t$, we have $N$ pedestrians. The 2D position of pedestrian $i$ at time step $t$ is denoted as $s_i^t \in \mathbb{R}^2$. We then can represent the past trajectories for all pedestrians in the past $H$ time steps as $x=s_{1,..,N}^{t-H+1:t} \in \mathbb{R}^{N \times H \times 2}$ and the respective future trajectories in the next $F$ time steps as $y=s_{1,..,N}^{t+1:t+F} \in \mathbb{R}^{N \times F \times 2}$. MATRIX is trained to model the distribution of $\mathbb{P}(y\mid x)$. To achieve the objective of realism, the behaviors generated by MATRIX shall match that of the real human behaviors recorded in the social pedestrian trajectory datasets. In the following, we present how MATRIX takes into consideration real data matching as well as the inductive biases for multi-modal properties. Fig. \ref{fig:model} illustrates the full architecture of MATRIX.

\subsection{History-and-Interaction-Aware Context Extraction}
To encode the highly dynamic and interactive social navigation scene, we adapt the encoder of Trajectron++ \cite{salzmann2020trajectron++}, including one node history LSTM and one edge history LSTM, denoted as $f$, to extract the rich and interactive context information from the multi-agent scene. We then represent the encoded history of all past trajectories as $e=f(x)$.

\subsection{Explicit Latent Variable Modeling for Multi-Future Destination Reasoning}

Temporal traveling destination is one of the most significant factors that affect human behavior. However, although in real crowd navigation scenes one can query the ground-truth future position as the temporal traveling destination, in data generation, the human shall exhibit multi-modal behavior. Therefore, in MATRIX, the first inference stream is to use an explicit latent stochastic model to capture this property.

Since the human potential destinations naturally follow the multi-kernel pattern, we assume joint normal distribution and design a two-dimensional Gaussian mixture model (GMM) to capture the temporal destination distribution. Concretely, the probability density function of temporal destination for each agent $i$ is $d_i \in \mathbb{R}^2$ is written as
\begin{equation}
P(d_i) = \sum_{k=1}^{K} c^k_i \cdot \frac{e^{ - \frac{1}{2} (d_i - \mu^k_i) (\Sigma^k_i)^{-1} (d_i - \mu^k_i)}}{{2\pi \cdot \sqrt{|\Sigma^k_i|}}},
\end{equation}
where $c^k_i \in \mathbb{R},\mu^k_i \in \mathbb{R}^2, \Sigma^k_i \in \mathbb{R}^{2\times 2}$ are the weight, mean, and covariance matrices for the $k$th normal distribution and $K$ is the number of Gaussian kernels. 
To obtain each of these quantities, we parameterize the encoded history with four fully connected layers $\{g_i\}_{i=1,2,3,4}$, each of which outputs the weights, means, log variances, and correlations of Gaussian kernels. 
In other words, we can rewrite the temporal destination inference stream as
\begin{equation}
    \begin{gathered}
        \pi^{1:K}_i = g_1(e), \mu^{1:K}_i = g_2(e), \\
\sigma^{1:K}_i= \exp{(g_3(e))},
\rho^{1:K}_i=\tanh{(g_4(e))}.
\end{gathered}
\end{equation}
We then reconstruct the covariance matrix of the $k$th component from the variance $\sigma^k_i$ and the correlation $\rho^k_i$ by
\begin{equation}
    \begin{gathered}
         \begin{bmatrix}
            \sigma_x \\ \sigma_y
        \end{bmatrix} = \sigma^k_i,
        \rho_{xy} =\rho^k_i,\\
        \Sigma^k_i = \begin{bmatrix}\sigma_x^2 & \rho_{xy}\sigma_x\sigma_y \\ \rho_{xy}\sigma_x\sigma_y & \sigma_y^2\end{bmatrix}.
    \end{gathered}
\end{equation}
We select $K$ to be large enough to capture the multi-modality. To encourage MATRIX to generate trajectories that match the real data, we employ an objective of maximum log-likelihood for the ground-truth temporal target $\bar{d}_i$ in the training objective:
\begin{equation}
\mathscr{L}_{\text{destination}} = - \frac{1}{N}\sum_{i=1}^{N} \log P(\bar{d}_i).
\end{equation}
In practice, since the training data is sparse and cannot represent the diverse potential future that MATRIX should be able to generate, the learned GMM can suffer from the mode collapse problem. 
We therefore apply a regularization on the weight, variance of the kernels, and distances between their centers. The resulting auxiliary loss is the summation of a series of hinge losses, which is written as
\begin{multline}
\mathscr{L}_{\text{mode}\_\text{collapse}} = \sum_{i=1}^{N} \sum_{k_1\neq k_2} \alpha_1 h(1-\beta_1 \|\mu^{k_1}_i-\mu^{k_2}_i\|_2)  \\+  \sum_{i=1}^{N}\sum_{k=1}^{K} \alpha_2 h(\beta_2 c^k_i-1) + \alpha_3 h(\beta_3 \|\sigma^k_i\|_2-1),
\end{multline}
where $\alpha$'s and $\beta$'s are hyperparameters and $h(\cdot)$ is the hinge loss function defined as
\begin{equation}
h(x) = \max(0, x).
\end{equation}
In the training time, we use the final position of the ground-truth trajectory $\bar{d}_i$ for the downstream decoder output. In the testing time, we sample diverse destinations $\hat{d}_i$ from GMMs and generate corresponding future motions with the decoder.

\subsection{Generation of Residual Actions}
After obtaining the latent representation of the spatio-temporal graph and temporal destination, the future motion is inferred using a Gated Recurrent Unit (GRU) decoder. The GRU cell is denoted as $g(\cdot)$ and the residual layer is denoted as $q(\cdot)$. Through the residual connection, MATRIX could autoregressively output the predicted trajectory by
\begin{equation}
    \begin{gathered}
        h_i^t = \begin{cases}g([e;\bar{d}_i;\hat{s}_i^t]) \text{ in training} \\ g([e;\hat{d}_i;\hat{s}_i^t]) \text{ in testing}\end{cases}
        \\\Delta\hat{s}_i^t= q(h_i^t),\hat{s}_i^{t+1} = \hat{s}_i^t +\Delta\hat{s}_i^t,
    \end{gathered}
\end{equation}
where $e$ is the encoded history of all the past trajectories, $\bar{d}_i$ is the ground truth temporal target, $\hat{d}_i$ is the sampled temporal destination from the GMM,  $h_i^t$ is the hidden state of GRU at time step $t$, $\Delta\hat{s}_i^t$ is the residual displacement, and $\hat{s}_i^t$ is the predicted state for agent $i$ at time step $t$.
\begin{table*}[t]
\small
\caption{\textbf{ASD/ADE/FDE values of 20 samples on the ETH/UCY dataset.} Bold indicates best.}
\vspace{-0.2cm}
\label{table1}
\begin{center}
\resizebox{\textwidth}{!}{\begin{tabular}{c|cbdcbdcbdcbdcbd}
\hline
\hline
\rowcolor{white}
\multirow{2}{*}{}&  \multicolumn{3}{c}{\textbf{UNIV}}& \multicolumn{3}{c}{\textbf{HOTEL}} & \multicolumn{3}{c}{\textbf{ZARA1}} &\multicolumn{3}{c}{\textbf{ZARA2}}&\multicolumn{3}{c}{\textbf{ETH}}\\
\hhline{~---------------}
\rowcolor{white}
&\textbf{ASD}&\textbf{ADE}&\textbf{FDE}&\textbf{ASD}&\textbf{ADE}&\textbf{FDE}&\textbf{ASD}&\textbf{ADE}&\textbf{FDE}&\textbf{ASD}&\textbf{ADE}&\textbf{FDE}&\textbf{ASD}&\textbf{ADE}&\textbf{FDE}\\
\hline
MID\cite{gu2022stochastic}
&N/A&0.22&0.45
&N/A&0.13&0.22
&N/A&0.17&0.30
&N/A&0.13&0.27
&N/A&0.39&0.66\\
PECNet\cite{mangalam2020pecnet}
&N/A&0.35&0.60
&N/A&0.18&0.24
&N/A&0.22&0.39
&N/A&0.17&0.30
&N/A&0.54&0.87\\
Y-Net\cite{mangalam2021goals}
&N/A&0.24&0.41
&N/A&0.10&\textbf{0.14}
&N/A&0.17&\textbf{0.27}
&N/A&0.13&\textbf{0.22}
&N/A&\textbf{0.28}&\textbf{0.33}\\
Trajectron++\cite{salzmann2020trajectron++}
&1.38&0.22&0.42
&1.12&0.12&0.19
&1.31&0.17&0.32
&1.08&\textbf{0.12}&0.25
&1.71&0.44&0.85\\
AgentFormer\cite{yuan2021agentformer}
&0.13&0.25&0.45
&1.00&0.14&0.22
&0.68&0.18&0.30
&0.44&0.14&0.24
&2.76&0.45&0.74\\
Social Implicit\cite{mohamed2020social}
&1.26&0.31&0.60
&2.39&0.20&0.36
&1.08&0.26&0.51
&1.20&0.22&0.43
&1.30&0.67&1.47\\
ExpertTraj+GMM\cite{he2021where}
&0.21&\textbf{0.19}&0.44
&0.11&\textbf{0.09}&0.15
&0.21&\textbf{0.15}&0.31
&0.14&\textbf{0.12}&0.24
&0.48&\textbf{0.30}&0.62\\
\hline
\textbf{MATRIX}
&\textbf{2.72}&0.22&\textbf{0.39}
&\textbf{2.87}&0.19&0.29
&\textbf{2.86}&0.20&0.35
&\textbf{2.41}&0.15&0.27
&\textbf{3.27}&0.94&1.61\\
\hline
\hline
\end{tabular}}
\end{center}
\vspace{-1.5em}
\end{table*}
\begin{table*}[t]
\small
\caption{\textbf{$\bm{\chi^2}$ distances on the ETH/UCY dataset.} Bold indicates best.}
\vspace{-0.2cm}
\label{table2}
\begin{center}
\begin{tabular}{c|cabd}
\hline
\hline
\rowcolor{white}
 Generated Data&Velocity &Acceleration &Angular Velocity &Angular Acceleration\\
 \hline
Imitation Learning Data
&1.740
&2.615
&0.103
&0.004\\
Trajectron++ Data
&\textbf{0.129}
&0.889
&0.071
&\textbf{0.002}\\
Agentformer Data
&0.645
&1.226
&0.025
&0.008\\
\hline
\textbf{MATRIX Data}
&0.184
&\textbf{0.763}
&\textbf{0.016}
&\textbf{0.002}\\
\hline
\hline
\end{tabular}
\end{center}
\end{table*}

To ensure MATRIX predicts correct motions, the sequential network is optimized by minimizing the Huber error between the predicted trajectory $\hat{s}_i^{\tau}$ and the ground-truth trajectory $s_i^{\tau}$. The reconstruction objective is written as
\begin{equation}
\mathscr{L}_{\text{reconstruction}} = \frac{1}{NF}\sum_{i=1}^{N} \sum_{\tau=t+1}^{t+F} L_{\text{Huber}}(\hat{s}^\tau_i, s^\tau_i).
\end{equation}
Overall, we train the network to minimize the combined loss
\begin{equation}
    \mathscr{L} = \lambda_1\mathscr{L}_{\text{destination}} + \lambda_2\mathscr{L}_{\text{mode}\_\text{collapse}} + \lambda_3\mathscr{L}_{\text{reconstruction}},
\end{equation}
where $\lambda_1, \lambda_2,$ and $\lambda_3$ are hyperparameters.

\section{Experimental Evaluation}
To demonstrate the effectiveness of MATRIX, we first illustrate the setup of the experiments. Then we describe a series of metrics that we selected and designed to evaluate the performance of MATRIX both as a predictor and a generator. Finally, we train an imitation-based motion planner on synthetic data generated by MATRIX and demonstrate the diversity of our generated results as well as the regularization effects of MATRIX data.

\subsection{Experiment Setup}
\subsubsection{Data Preparation}
MATRIX is trained on two widely used datasets: the ETH dataset \cite{pellegrini2009you}, with subsets named ETH and HOTEL, and the UCY dataset \cite{lerner2007crowds}, with subsets named ZARA1, ZARA2, and UNIV. Both datasets provide interactive human pedestrian navigation episodes and provide key information on pedestrian position and velocity. In our experiments, we use the real initial state from the dataset as initialization. In the training phase, the data is segmented into batches that consist of observed trajectories of 8 time steps, each of which corresponds to 3.2 seconds, and future trajectories of the next 12 time steps, which corresponds to 4.8 seconds. We follow the leave one out cross-validation as previous works \cite{gupta2018social, salzmann2020trajectron++}.

\subsubsection{Implementation Details}

We designed four fully connected layers to model the weights, means, variances, and covariances of GMM. We set $K$ the number of GMM components to 4. For the decoder, we employ a Gated Recurrent Unit (GRU), whose hidden size is 128, and a linear layer to output the residual action. 
MATRIX is implemented with PyTorch and trained with Intel Core I7 CPUs and NVIDIA RTX 2080 Ti GPUs for 100 epochs. The training iterations take a data batch of size 256. The learning rate is set to $0.001$ initially and decays exponentially every epoch with a decay rate of 0.9999. The model is trained with Adam optimizer and gradients are clipped at $1.0$.

\subsubsection{Evaluation Metrics}

We compare MATRIX against baselines based on a series of metrics demonstrating the diversity and realism properties of the generated trajectories. Above all, we evaluate how well MATRIX produces diverse and realistic contexts. Therefore, we include:
\paragraph{Diversity (ASD)} The diversity of the generated data is an important metric. We adopt the average self distance (ASD) \cite{yuan2019diverse} to measure how the generator can produce diverse contexts. To compute the ASD, we generate samples and filter out the trajectories with any collision, resulting in 20 samples. Then, we find the maximum of the average distances across time between any of the two samples $s_{l_1,i}^\tau$ and $s_{l_2,i}^\tau$:
\begin{equation}
\text{ASD} = \frac{1}{F}\max_{l_1, l_2 \in \{1,\cdots,20\}} \sum_{\tau=t+1}^{t+F} \|s_{l_1,i}^\tau-s_{l_2,i}^\tau \|_2,
\end{equation}

We follow previous trajectory reconstruction literature and report the distance error metrics as a proxy of our model's capability of producing realistic trajectories \cite{alahi2016social,salzmann2020trajectron++}:

\begin{figure*}[!tbp]
\hspace{0.4cm}{\large ETH} \hspace{2.2cm}{\large HOTEL}\hspace{2.2cm} {\large UNIV}\hspace{2.2cm} {\large ZARA1}\hspace{2.2cm}{\large ZARA2}
\vskip 0.04cm
    \centering 
\begin{subfigure}{0.2\textwidth}
  \includegraphics[width=0.99\linewidth, height=2.81cm]{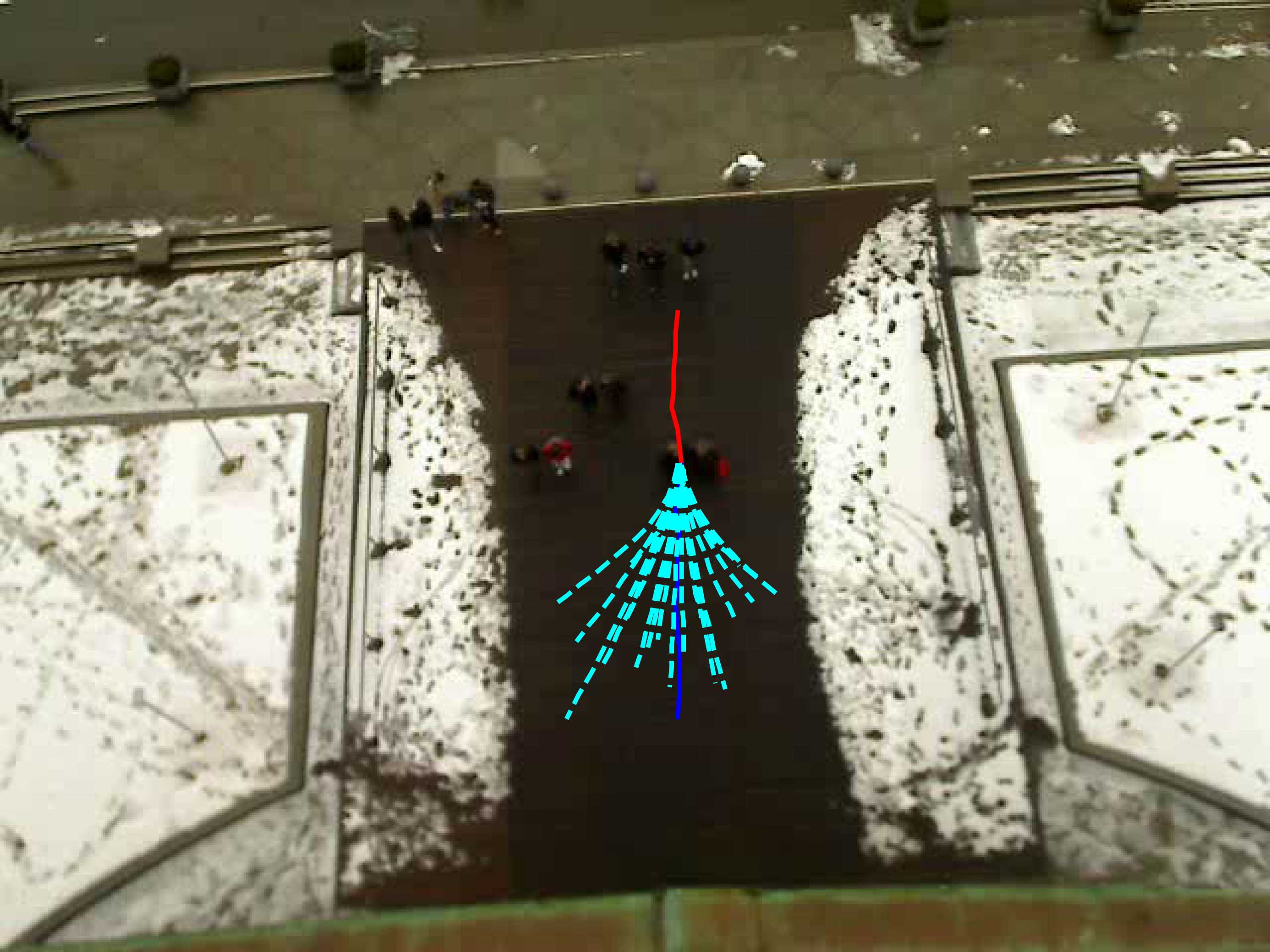}
\end{subfigure}\hfil 
\begin{subfigure}{0.2\textwidth}
  \includegraphics[width=0.99\linewidth]{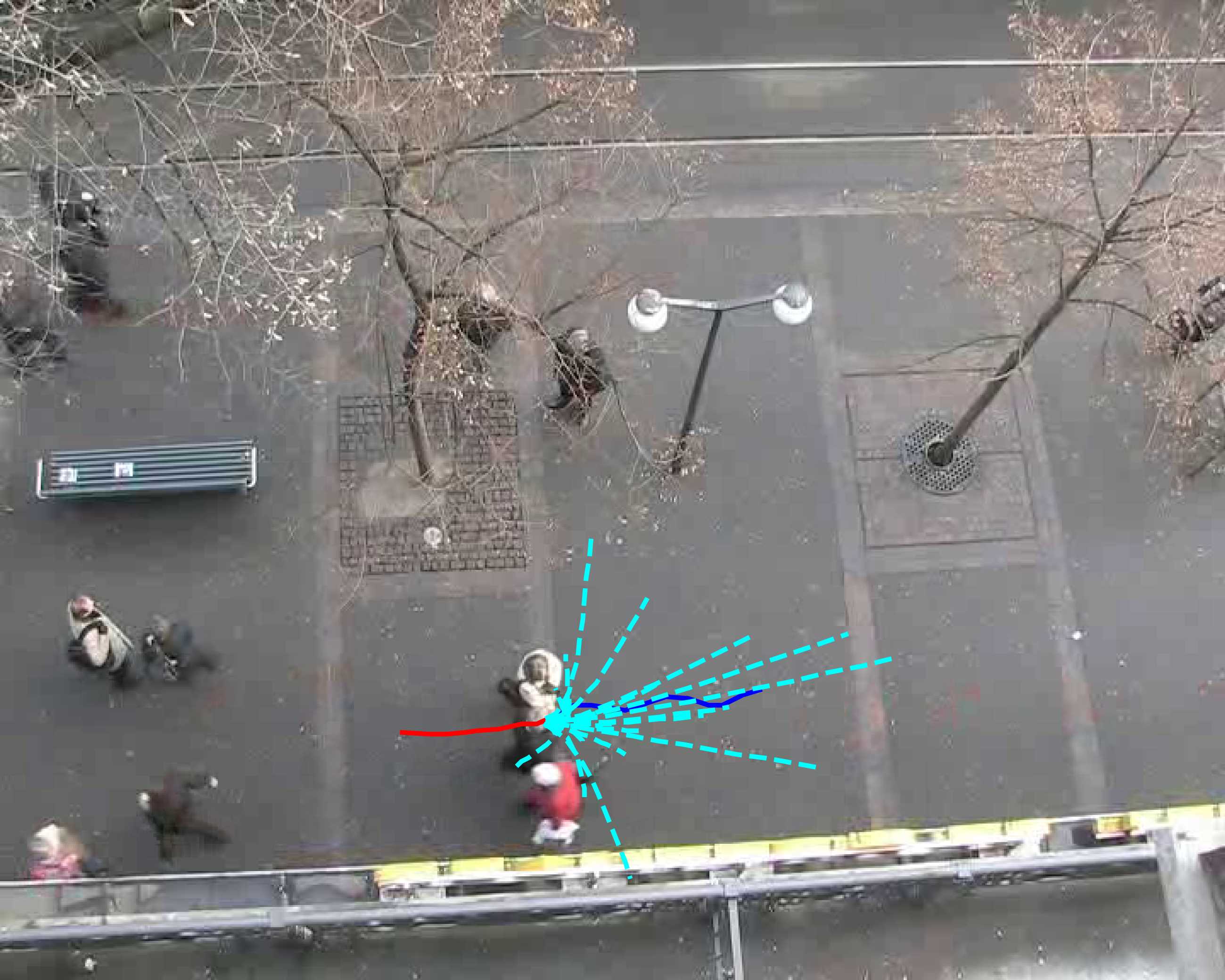}
\end{subfigure}\hfil 
\begin{subfigure}{0.2\textwidth}
  \includegraphics[width=0.99\linewidth]{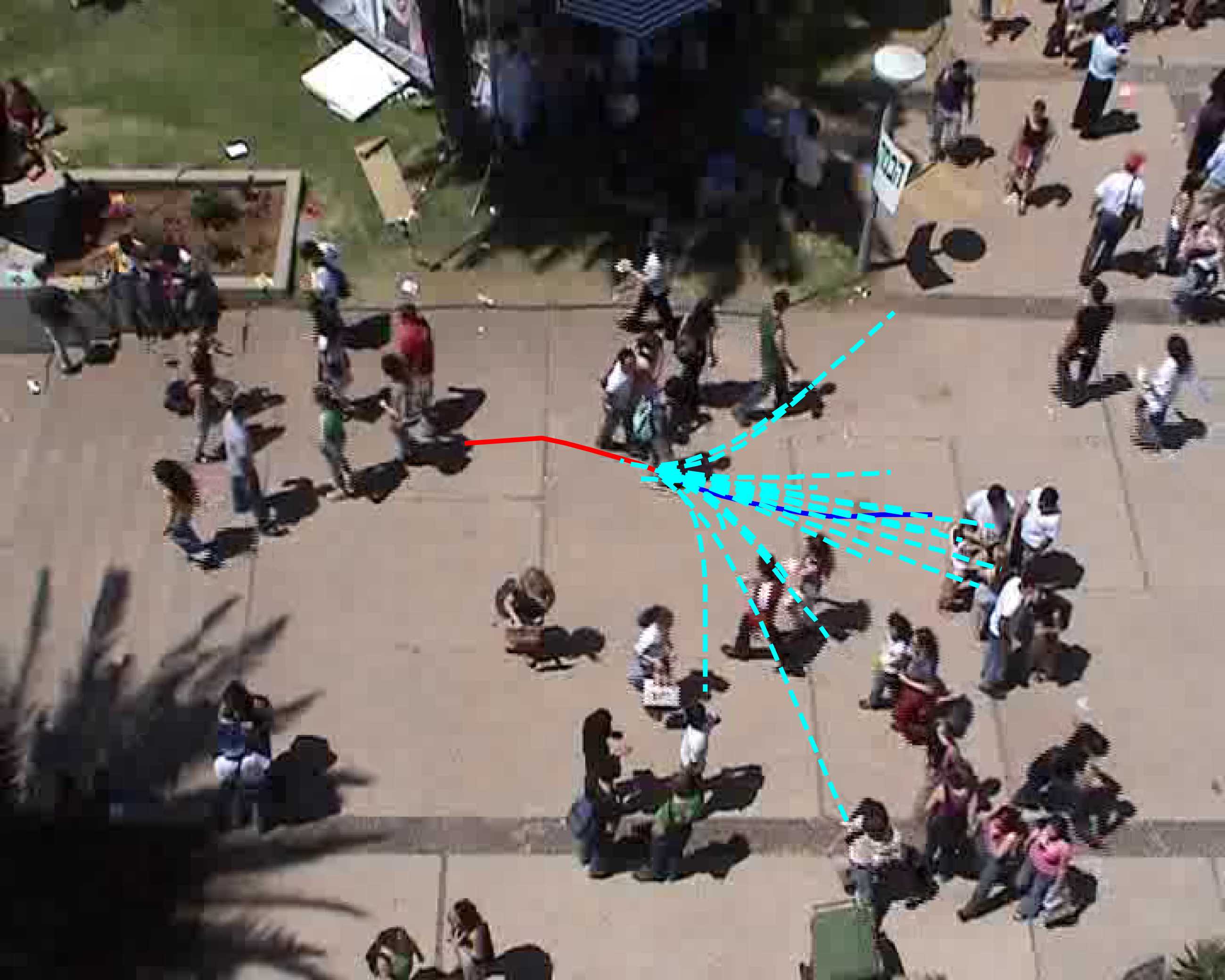}
\end{subfigure}\hfil 
\begin{subfigure}{0.2\textwidth}
  \includegraphics[width=0.99\linewidth]{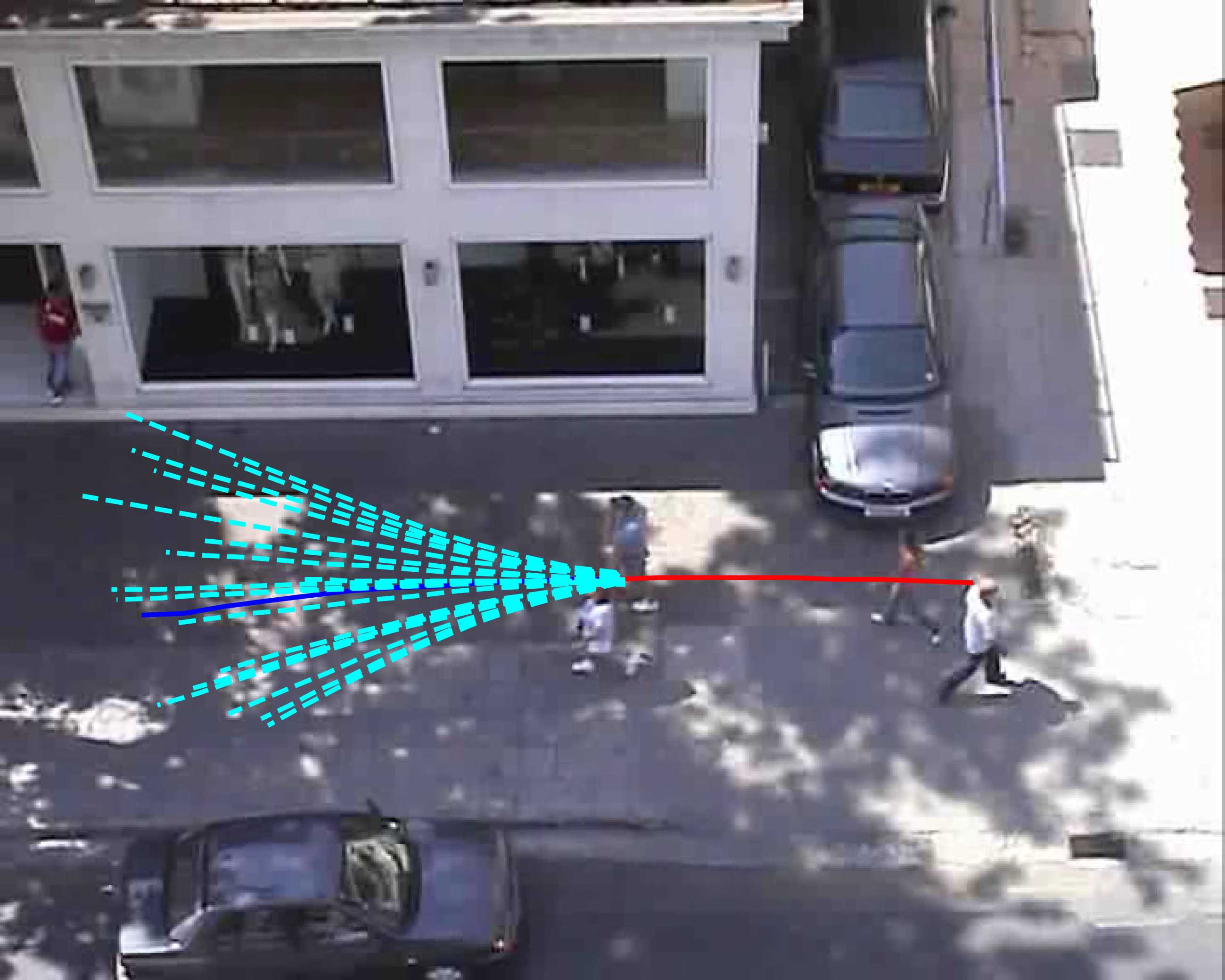}
\end{subfigure}\hfil 
\begin{subfigure}{0.2\textwidth}
  \includegraphics[width=0.99\linewidth]{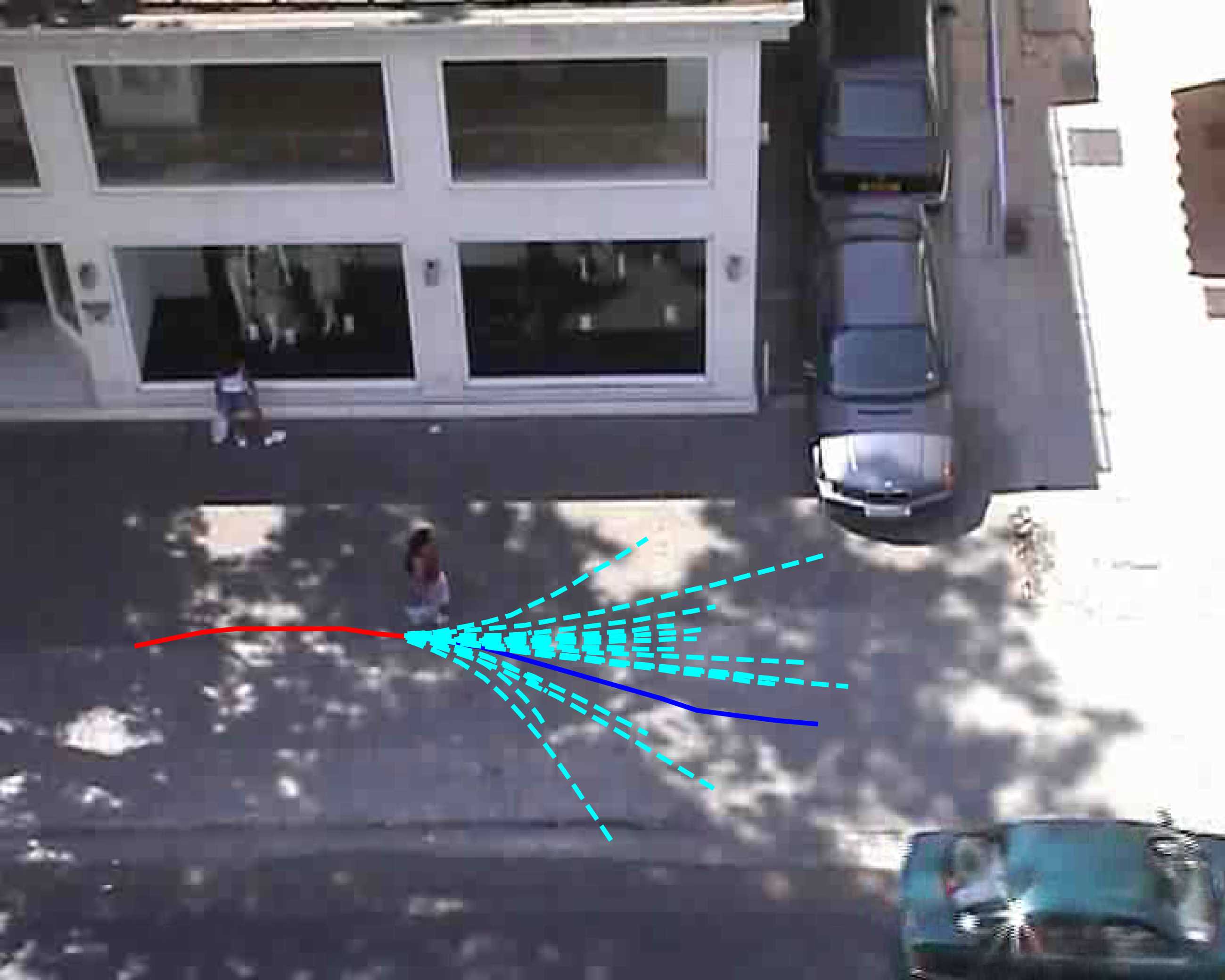}
\end{subfigure}
\vskip 0.04cm
\begin{subfigure}{0.2\textwidth}
  \includegraphics[width=0.99\linewidth, height=2.81cm]{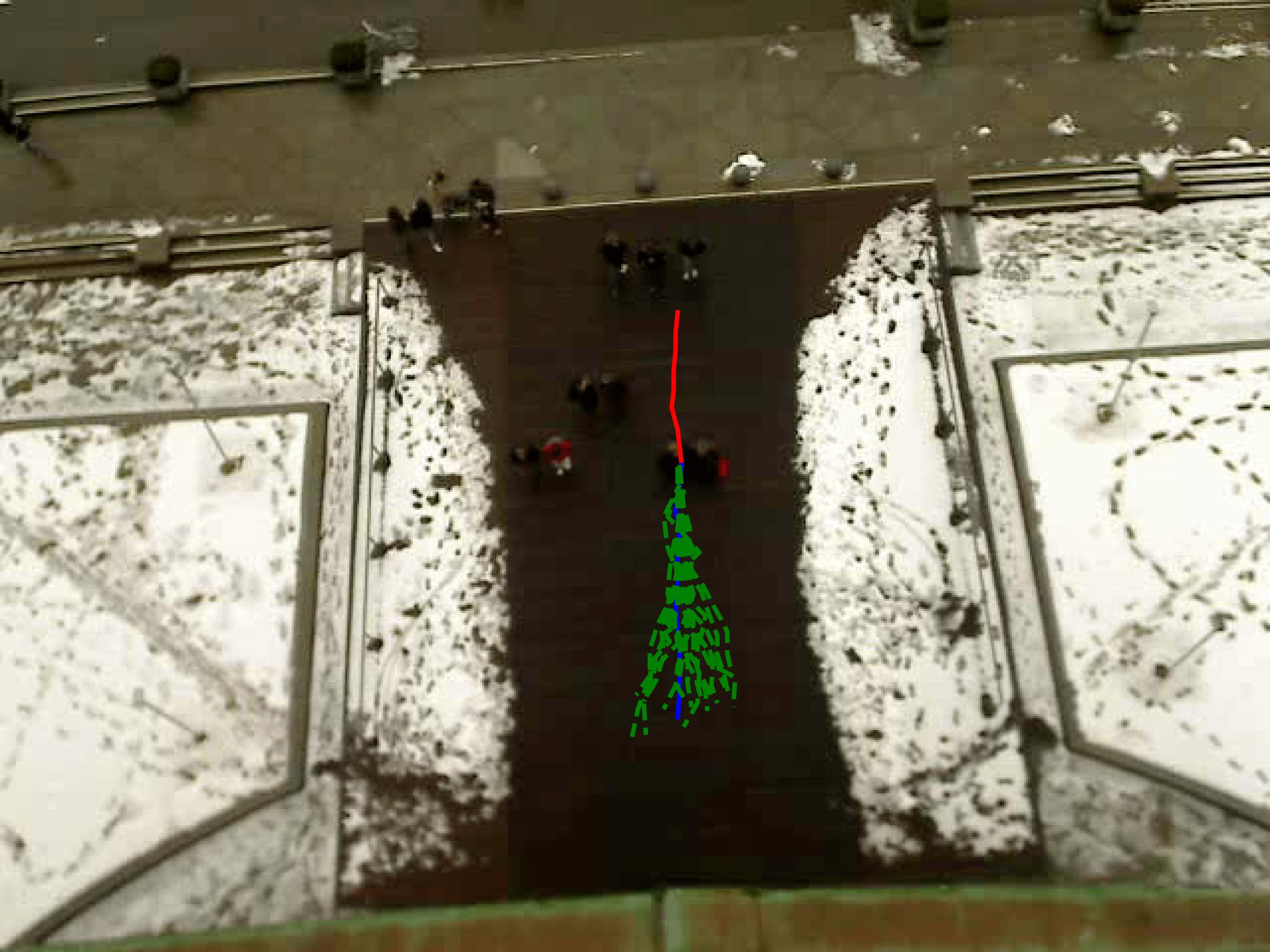}
\end{subfigure}\hfil 
\begin{subfigure}{0.2\textwidth}
  \includegraphics[width=0.99\linewidth]{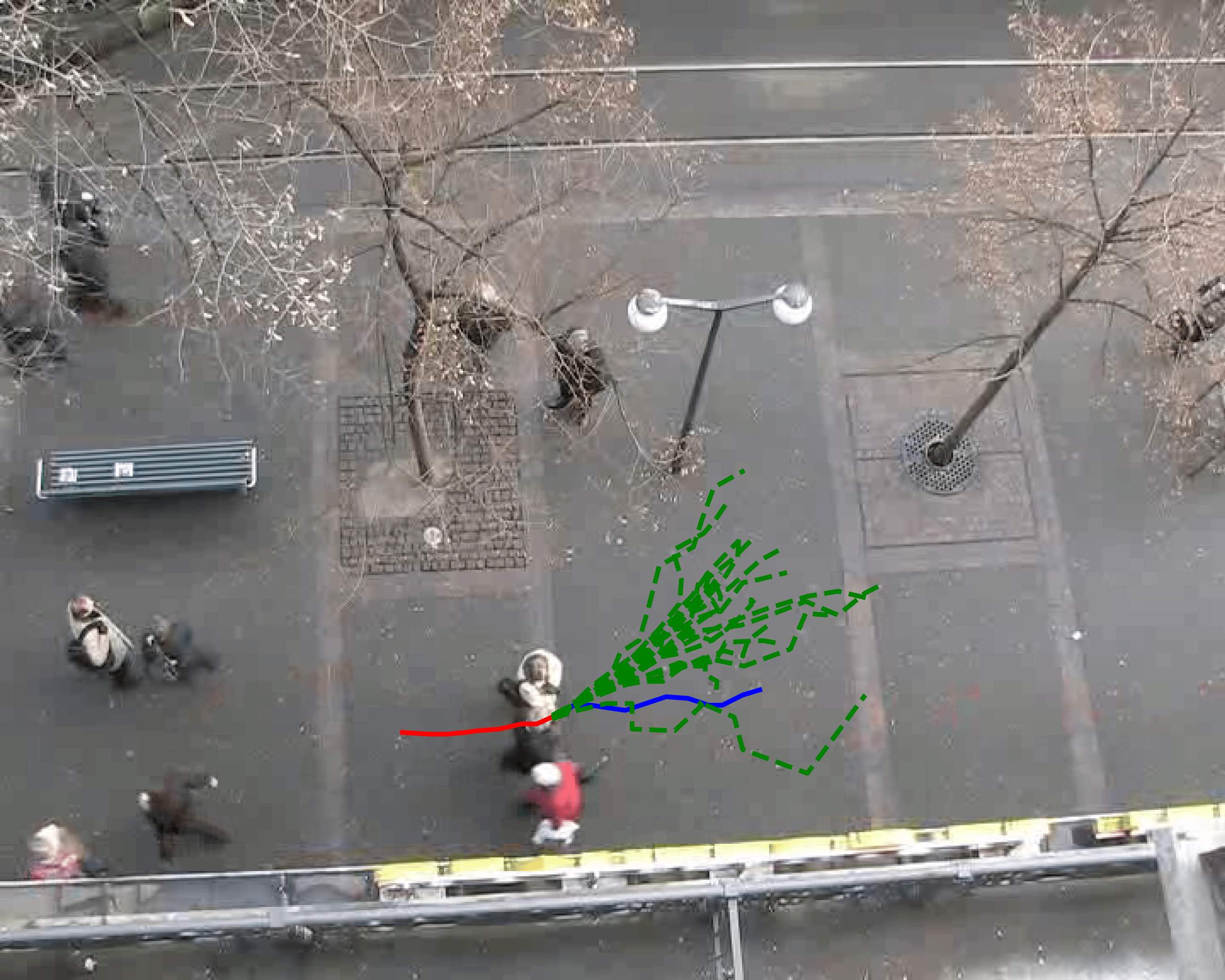}
\end{subfigure}\hfil 
\begin{subfigure}{0.2\textwidth}
  \includegraphics[width=0.99\linewidth]{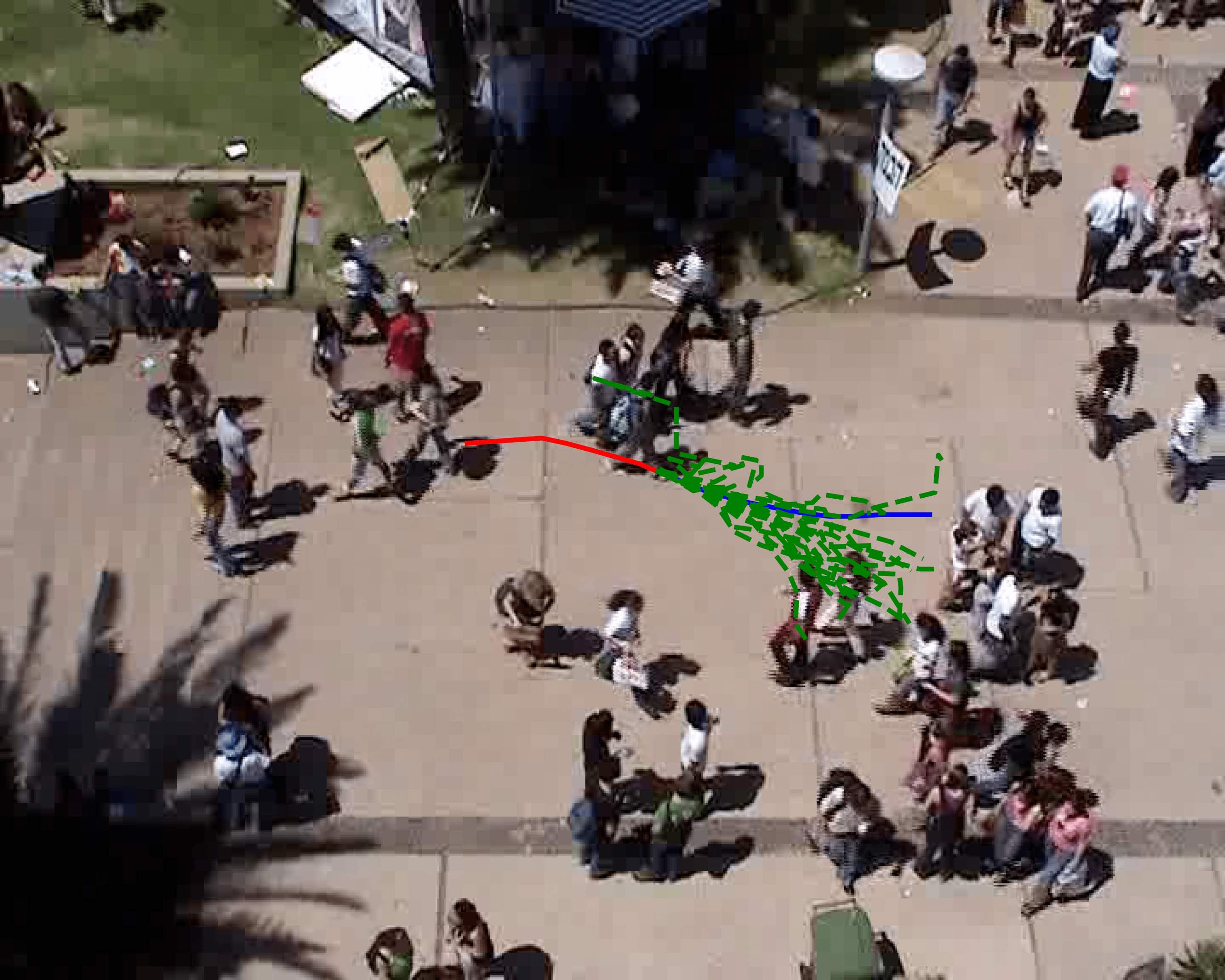}
\end{subfigure}\hfil 
\begin{subfigure}{0.2\textwidth}
  \includegraphics[width=0.99\linewidth]{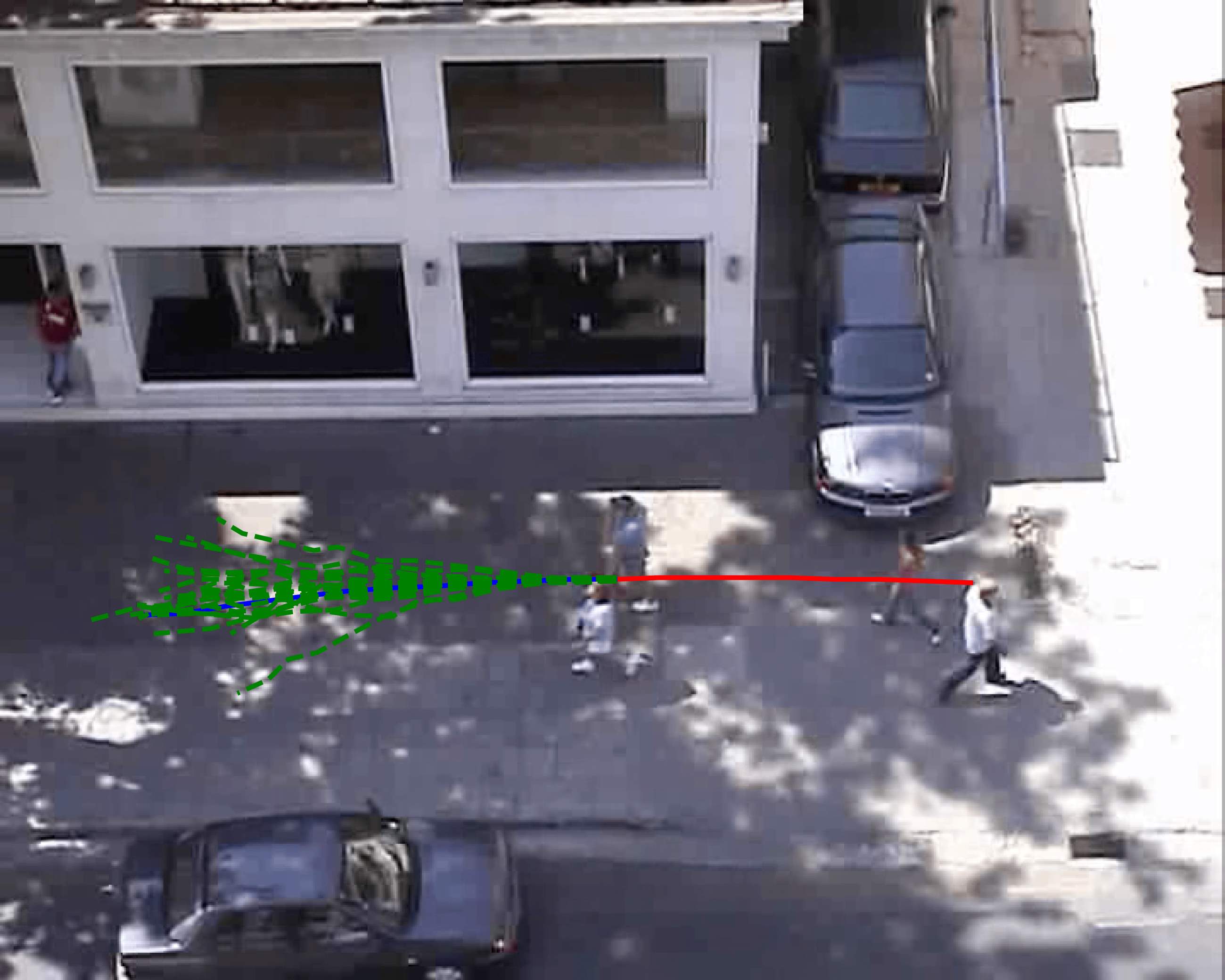}
\end{subfigure}\hfil 
\begin{subfigure}{0.2\textwidth}
  \includegraphics[width=0.99\linewidth]{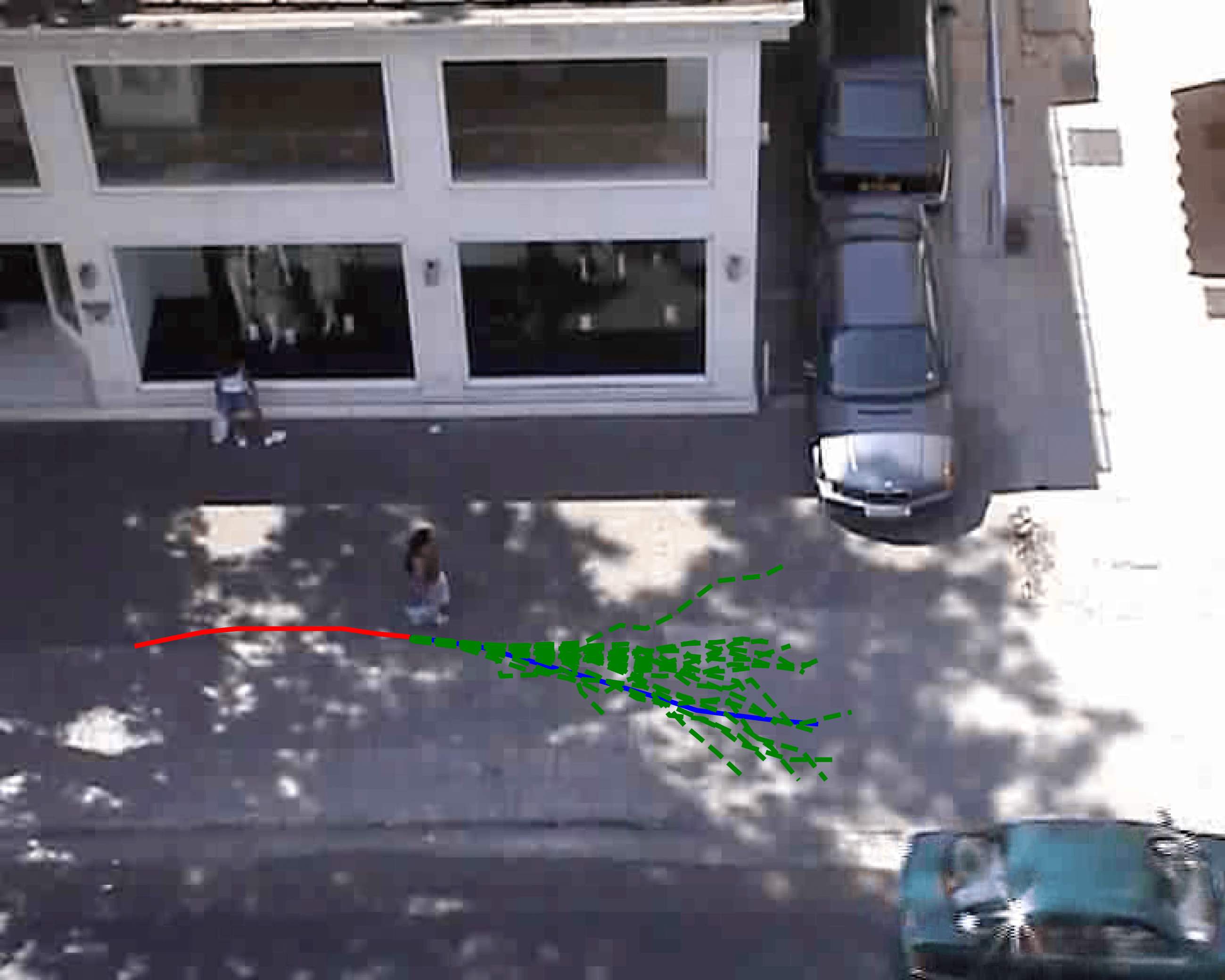}
\end{subfigure}
\vskip 0.04cm
Past Trajectory \textcolor{red}{\textbf{---------------}}\hspace{0.25cm} Future Trajectory \textcolor{blue}{\textbf{---------------}}\hspace{0.25cm} MATRIX \textcolor{cyan}{\textbf{-- -- -- -- -- -- --}}\hspace{0.25cm} Trajectron++ \textcolor[rgb]{0.0, 0.5, 0.0}{\textbf{-- -- -- -- -- -- --}}\hspace{0.25cm}
\caption{\textbf{Visualization of generated trajectories.} Provided with the past trajectory (red), MATRIX (cyan) and Trajectron++ (green) can generate 20 possible future trajectories for five different scenes. We see that our generated trajectories are much more diverse than Trajectron++. Zoom in for better visualization.}
\label{fig:diversity}
\end{figure*}
\begin{figure*}[!tbp]
\vspace{-1.em}
    \centering 
\begin{subfigure}{0.2\textwidth}
  \includegraphics[width=0.99\linewidth, height=2.81cm]{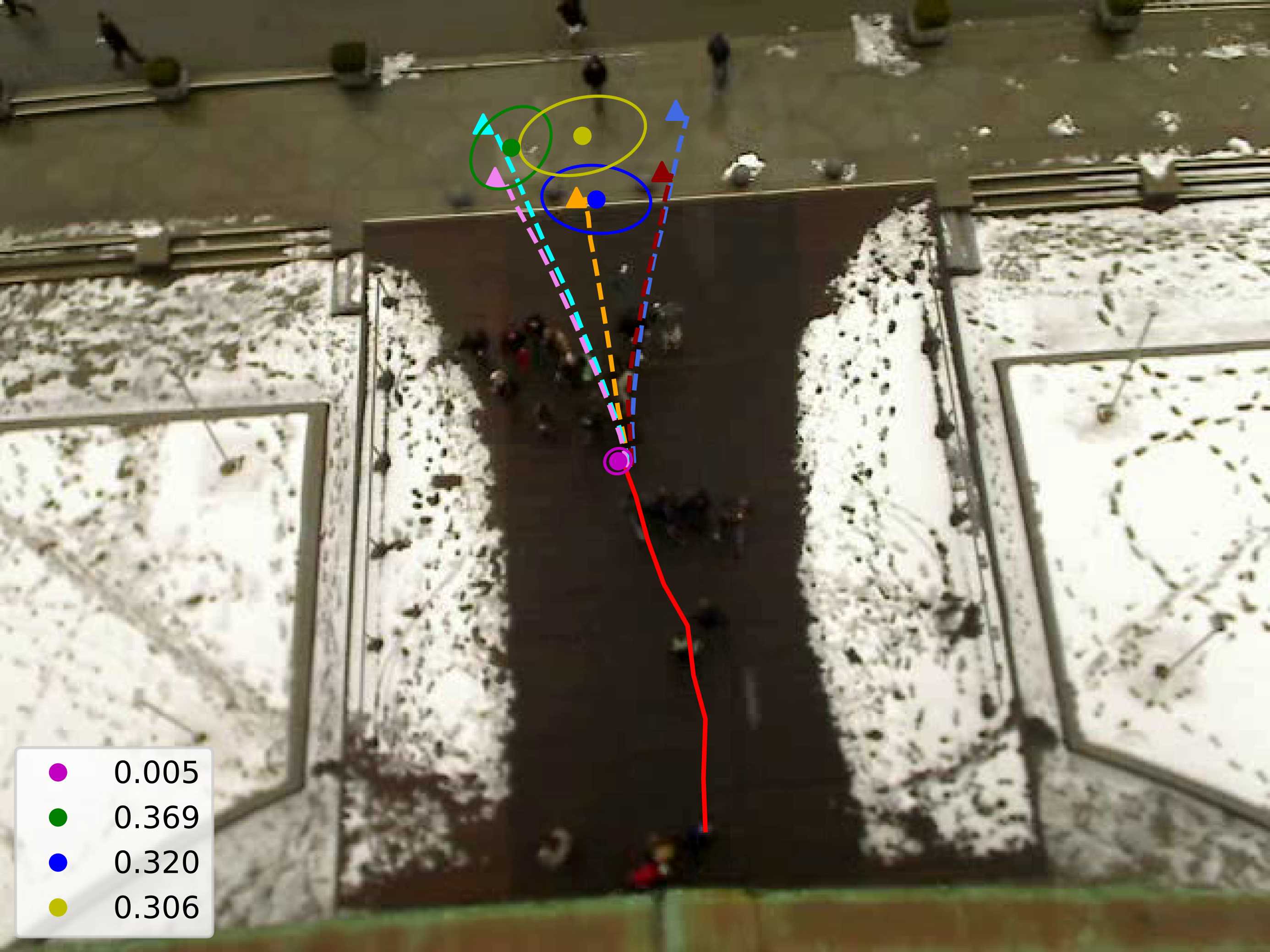}
\end{subfigure}\hfil 
\begin{subfigure}{0.2\textwidth}
  \includegraphics[width=0.99\linewidth]{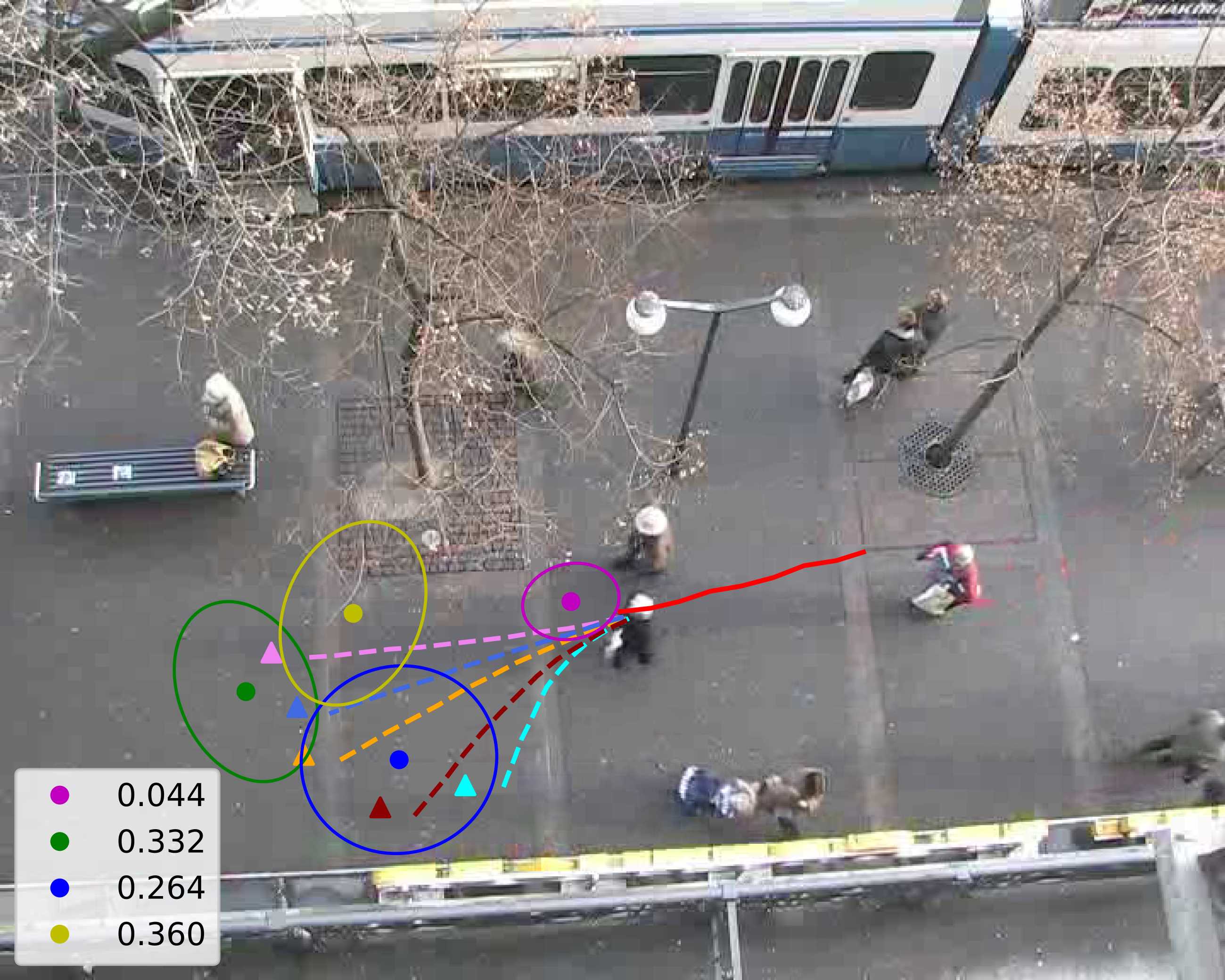}
\end{subfigure}\hfil 
\begin{subfigure}{0.2\textwidth}
  \includegraphics[width=0.99\linewidth]{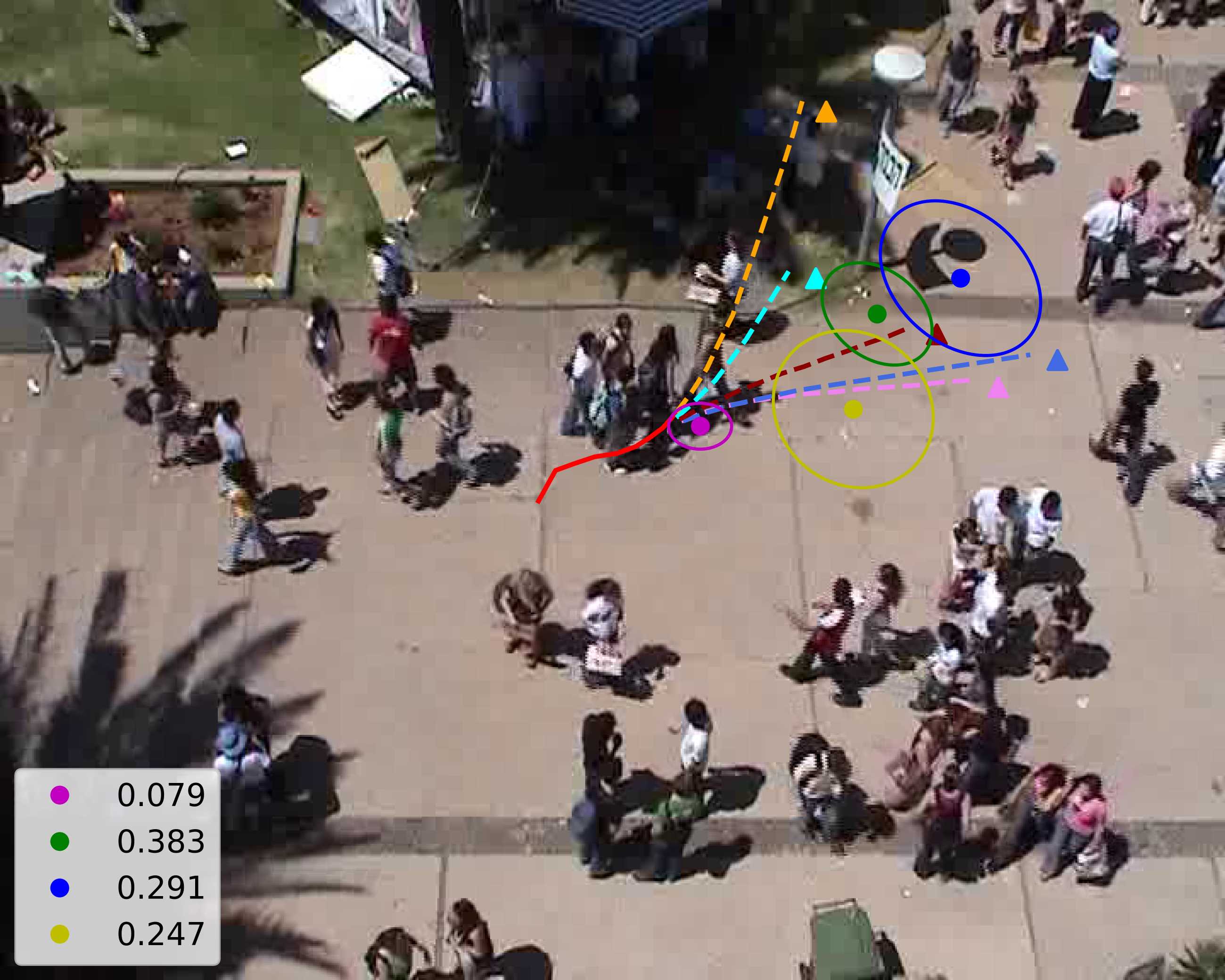}
\end{subfigure}\hfil 
\begin{subfigure}{0.2\textwidth}
  \includegraphics[width=0.99\linewidth]{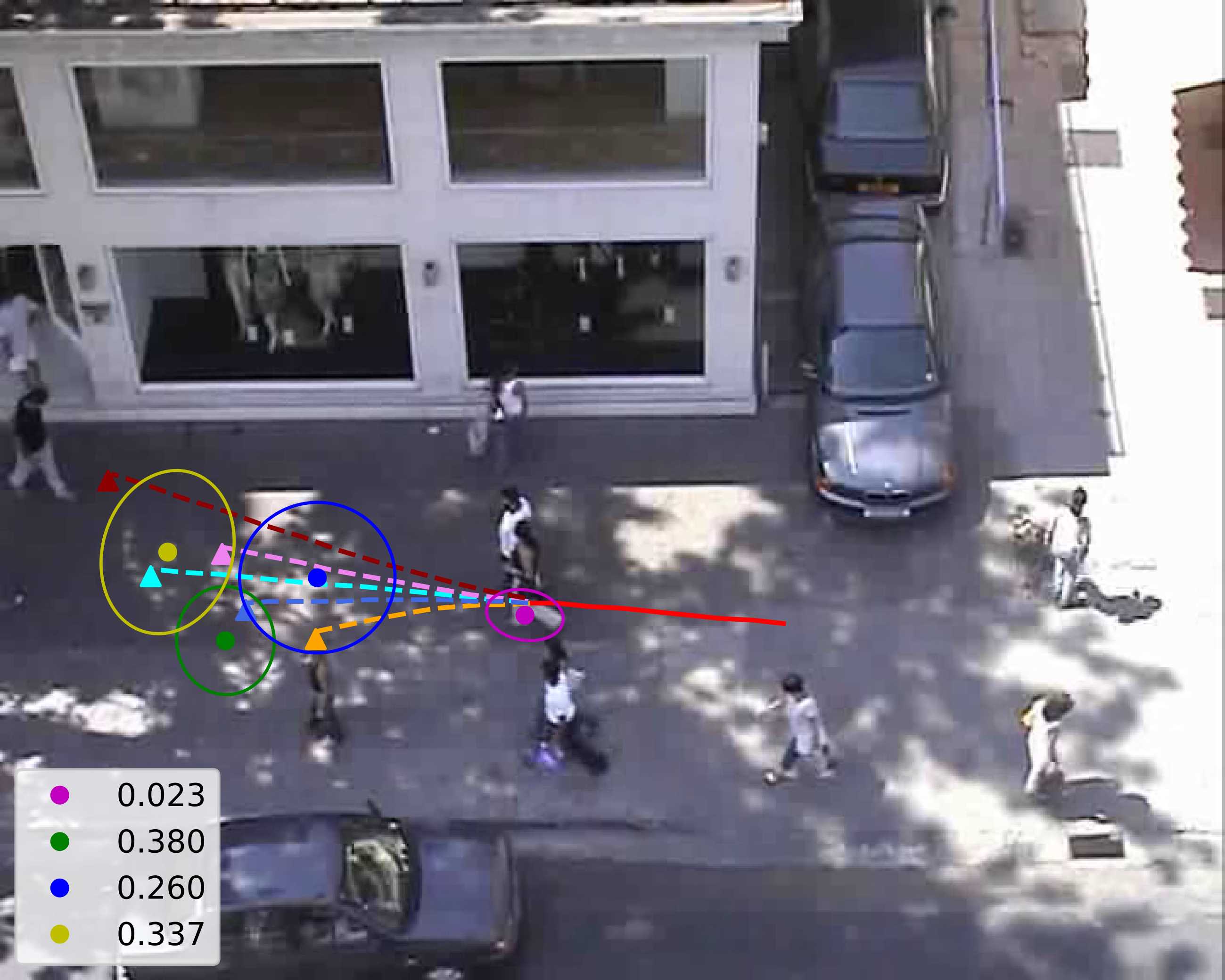}
\end{subfigure}\hfil 
\begin{subfigure}{0.2\textwidth}
  \includegraphics[width=0.99\linewidth]{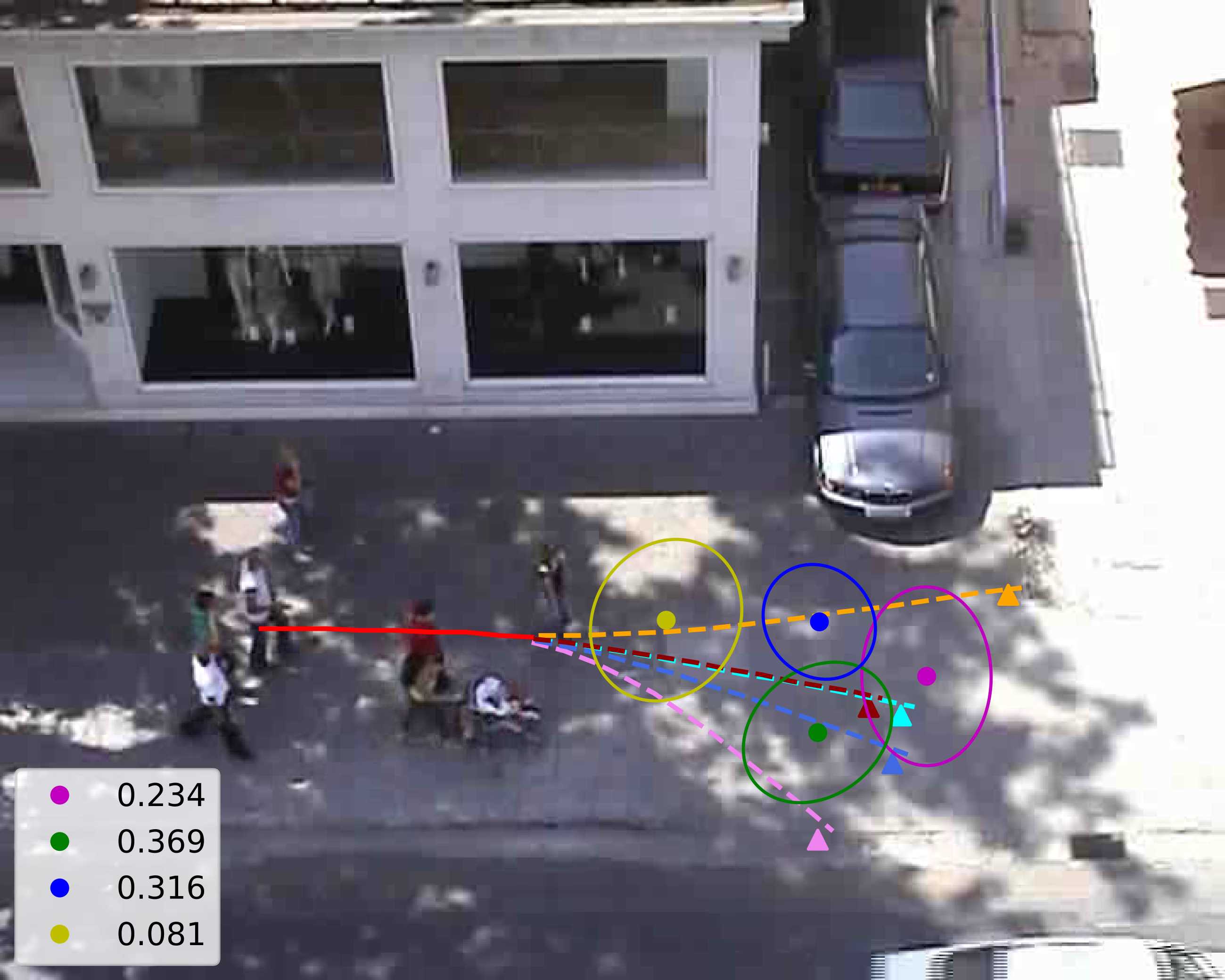}
\end{subfigure}
\caption{\textbf{Visualization of GMM.} We use MATRIX to generate five future trajectories (orange, cyan, violet, dark red, and royal blue) based on the past trajectory (red) and five stochastic destinations, represented as triangles, sampled from GMM. The center of each ellipse (green, magenta, yellow, and dark blue) is the mean of each Gaussian, and the radius is its one standard deviation. The weight of each Gaussian can be found in the legend. Zoom in for better visualization.}
\label{fig:gmm}
\end{figure*}
\paragraph{Average Displacement Error (ADE)} ADE is the mean $\textit{l}_2$ distance between the ground truth $s_i^\tau$ and predictions $\hat{s}_i^\tau$:
\begin{equation}
\text{ADE} = \frac{1}{N}\frac{1}{F}\sum_{i=1}^{N}\sum_{\tau=t+1}^{t+F} \|\hat{s}_i^\tau-s_i^\tau \|_2.
\end{equation}
\paragraph{Final Displacement Error (FDE)} FDE is the $\textit{l}_2$ distance between the predicted final position $s_i^{t+F}$ and the ground-truth final position $\hat{s}_i^{t+F}$:
\begin{equation}
    \text{FDE} = \frac{1}{N}\sum_{i=1}^{N} \| \hat{s}_i^{t+F}-s_i^{t+F}\|_2,
\end{equation}
where $N$ is the total number of pedestrians and $F$ is the number of future horizons. 

Since MATRIX is supposed to generate stochastic multi-modal contexts, we sample 20 trajectories and evaluate the best matching mode, and compute the best-of-20 ADE/FDE.
\paragraph{Chi-square Distance ($\bm{\chi^2}$)} The capability of reconstructing the training trajectories is no longer a fair metric for realism under the setting of trajectory generation, and thus we introduce a new set of measurements of the Chi-square distance between key motion primitive distributions of the generated data and the original data. We compare four primitives, including velocity, acceleration, angular velocity, and angular acceleration. 
\begin{equation}
    {\chi^2} = \sum_{i=1}^{20} \frac{(x_i-y_i)^2}{x_i+y_i},
\end{equation}
where $x_i$ and $y_i$ are the estimated probability density of the generated data and the raw data in the $i$th bin in a total of 20 bins. Since all five subsets come from similar scenarios, we compute the average Chi-square Distance over them.

\begin{table*}[t]
\small
\caption{\textbf{ASD/ADE/FDE values for ablation study.} $\mathbf{MC}$ = Mode Collapse, $\bigoplus$ = Residual Actions. Bold indicates best.}
\vspace{-0.2cm}
\label{table3}
\begin{center}
\resizebox{\textwidth}{!}{\begin{tabular}{cc|cbdcbdcbdcbdcbd}
\hline
\hline
\rowcolor{white}
\multirow{2}{*}{MC}&\multirow{2}{*}{$\bigoplus$}&  \multicolumn{3}{c}{\textbf{UNIV}}& \multicolumn{3}{c}{\textbf{HOTEL}} & \multicolumn{3}{c}{\textbf{ZARA1}} &\multicolumn{3}{c}{\textbf{ZARA2}}&\multicolumn{3}{c}{\textbf{ETH}}\\\hhline{~~|---------------}

&&\textbf{ASD}&\textbf{ADE}&\textbf{FDE}&\textbf{ASD}&\textbf{ADE}&\textbf{FDE}&\textbf{ASD}&\textbf{ADE}&\textbf{FDE}&\textbf{ASD}&\textbf{ADE}&\textbf{FDE}&\textbf{ASD}&\textbf{ADE}&\textbf{FDE}\\
\hline
-&-
&2.12&0.29&0.53
&2.56&0.24&0.41
&2.34&0.25&0.47
&1.82&0.20&0.36
&3.09&0.96&1.71
\\
-&\checkmark
&2.06&0.28&0.53
&2.34&0.23&0.39
&2.30&0.24&0.47
&1.85&0.19&0.37
&2.86&0.96&1.70\\
\checkmark&-
&2.34&0.28&0.52
&2.48&0.24&0.40
&2.50&0.26&0.48
&2.03&0.20&0.37
&2.78&1.09&1.89\\
\hline
\checkmark&\checkmark
&\textbf{2.72}&\textbf{0.22}&\textbf{0.39}
&\textbf{2.87}&\textbf{0.19}&\textbf{0.29}
&\textbf{2.86}&\textbf{0.20}&\textbf{0.35}
&\textbf{2.41}&\textbf{0.15}&\textbf{0.27}
&\textbf{3.27}&\textbf{0.94}&\textbf{1.61}\\
\hline
\hline
\end{tabular}}
\end{center}
\vspace{-1.5em}
\end{table*}
\begin{table*}[t]
\small
\caption{\textbf{ADE/FDE values for imitation learning with data augmentations.} Bold indicates best.}
\vspace{-0.2cm}
\label{table4}
\begin{center}
\begin{tabular}{c|cbcbcbcbcb}
\hline
\hline
\rowcolor{white}
\multirow{2}{*}{}& \multicolumn{2}{c}{\textbf{UNIV}} & \multicolumn{2}{c}{\textbf{HOTEL}}& \multicolumn{2}{c}{\textbf{ZARA1}} &\multicolumn{2}{c}{\textbf{ZARA2}}& \multicolumn{2}{c}{\textbf{ETH}}\\
\hhline{~----------}
&\textbf{ADE}&\textbf{FDE}&\textbf{ADE}&\textbf{FDE}&\textbf{ADE}&\textbf{FDE}&\textbf{ADE}&\textbf{FDE}&\textbf{ADE}&\textbf{FDE}\\
\hline
Raw Data
&0.38&\textbf{0.43}
&0.51&0.52
&0.27&0.36
&\textbf{0.21}&0.28
&\textbf{0.69}&\textbf{0.85}\\
Trajectron++ Data
&0.35&\textbf{0.43}
&0.46&0.70
&\textbf{0.25}&\textbf{0.35}
&\textbf{0.21}&0.30
&0.75&1.05\\
\hline
\textbf{MATRIX Data}
&\textbf{0.34}&\textbf{0.43}
&\textbf{0.44}&\textbf{0.48}
&\textbf{0.25}&\textbf{0.35}
&\textbf{0.21}&\textbf{0.27}
&0.71&0.91\\
\hline
\hline
\end{tabular}
\end{center}

\end{table*}
\subsection{Quantitative Comparisons}
We compare our model with a wide range of state-of-the-art models in Table \ref{table1}. Note that some of the models, such as Y-Net and ExpertTraj-GMM, achieve lower reconstruction errors at the cost of the diversity metric ASD. In contrast, MATRIX has a significantly higher ASD value. For example, the HOTEL dataset has an ASD score of 2.87, which is 100.7\% better than that of Trajectron++. Meanwhile, trajectories generated by MATRIX have ADE and FDE values with the same orders of magnitude compared with models specialized for trajectory reconstruction, demonstrating that MATRIX can maintain a relatively low reconstruction error even when the generated behaviors are far more diverse. 

\begin{figure}
\begin{center}
\includegraphics[width=\linewidth]{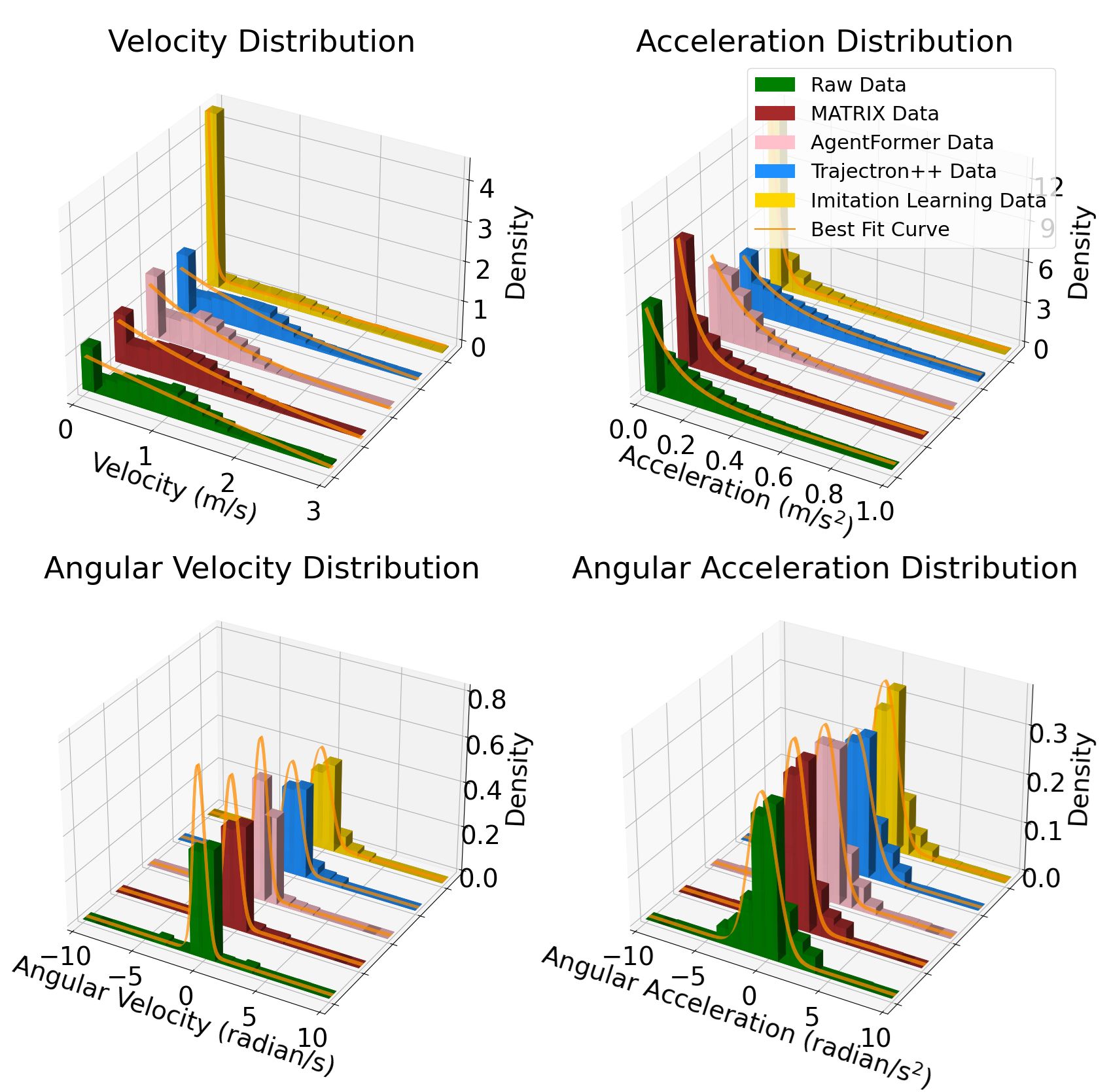}
\end{center}
\caption{
\textbf{Physics primitives of the generated data.} We plot the histogram of physics primitives of four generated datasets -- MATRIX data (red), Agentformer Data (pink), Trajectron++ Data (blue), and Imitation Learning Data (yellow) -- against the raw one (green). The orange line is the best-fit curve. Note that we use exponential distribution for velocity and acceleration and Gaussian distribution for angular velocity and angular acceleration.}
\vspace{-0.4cm}
\label{fig:distribution}
\end{figure}

Since reconstructing the training trajectories is not sufficient for realism evaluation under the circumstance of diverse trajectory generation. In Table \ref{table2} and Fig. \ref{fig:distribution}, we show how the distribution shifts of key motion primitives for trajectories generated by MATRIX compare with trajectories generated using other methods. MATRIX has the lowest $\chi^2$ scores across most motion primitives, indicating that the MATRIX data indeed matches the distribution of human motions. This is attributed to the learnable residual action, as in contrast, Trajectron++ has unlearnable dynamic integration via dynamics, making it much more difficult to control over the randomness of GMM and produce the distributions matching the real one \cite{salzmann2020trajectron++}.

\subsection{Multi-modality}
We further visualize the diverse trajectories generated by MATRIX. Fig. \ref{fig:diversity} illustrates the 20 sampled predictions on all five subsets. The qualitative results show that though both MATRIX and Trajectron++ \cite{salzmann2020trajectron++} can generate heterogeneous sequences of paths, MATRIX produces much more diverse outcomes. In addition, we observe that MATRIX's samples are much smoother than Trajectron++ and match the real behaviors of human motions. We deduce this feature is attributed to the advantage of employing both GMM and residual actions. Specifically, residual connection encourages each agent to move toward the stochastic destinations sampled from GMM and thus results in reasonable paths. To better understand how each of the two components works, we sample five different targets from GMM and visualize each respective trajectory in Fig. \ref{fig:gmm}. We can see each generated trajectory is exactly driven by the target, showing the effectiveness of using GMMs to model the explicit latent variable and residual action to control the direction.

\subsection{Ablation Study}
To demonstrate the significance of each component in MATRIX, we perform a comprehensive ablation study in Table \ref{table3}. We show that without the mode collapse loss, MATRIX suffers from higher reconstruction errors and lower diversity because each GMM collapses to a single point with unknown variance. In addition, removing the residual action scheme leads to an increment in both displacement errors.

\subsection{Serving as a Data Augmentation}
To further investigate the realism of the data generated by MATRIX, we use the samples generated by MATRIX as an augmentation dataset. We combine UCY and ETH datasets with synthetic data generated by MATRIX from the original datasets and use the combined dataset to train imitation learning planners. We evaluate the planner performance on the evaluation datasets against planners trained using the original datasets only and augmented with synthetic data generated by Trajectron++. All models were of the same structure of 10-layer and 128 hidden-size LSTM with residual layers, and trained for 170 epochs. The results show that the imitation learning model learned using data generated by MATRIX can produce lower reconstruction errors in Table \ref{table4}, which is significant since with zero extra data beyond the training data, MATRIX's generated data improved the imitation learning planner performance on unseen evaluation datasets. Hence, we conclude MATRIX can produce both diverse and realistic samples that are beneficial for downstream tasks.
\section{Conclusions}
In this paper, we introduce MATRIX, a data generator for multi-agent human trajectory generation with diverse contexts. By explicitly modeling the significant factor that affects heterogeneous human behaviors - the temporal destination - and controlling moving direction through a residual network, MATRIX generates multi-modal behaviors that realistically interact with external agents. Our experiments demonstrate the realism and diversity of the MATRIX data, as well as its potential to serve as a predictor.

{\small
\bibliographystyle{IEEEtran}
\bibliography{egbib,references}
}

\end{document}